David Azatyan

# Image Classification of Stroke Blood Clot Origin using Deep Convolutional Neural Networks and Visual Transformers: A Kaggle Competition

David Azatyan
dgazatyan@gmail.com

**Keywords:** Computer vision, Artificial Intelligence, Medical imaging, Image Classification, Deep learning

## Abstract

*Stroke is one of two main causes of death worldwide. Many individuals suffer from ischemic stroke every year. Only in US more over 700,000 individuals meet ischemic stroke due to blood clot blocking an artery to the brain every year.*

*«Stroke symptoms typically start suddenly, over seconds to minutes, and in most cases do not progress further. The symptoms depend on the area of the brain affected. The more extensive the area of the brain affected, the more functions that are likely to be lost. » [10]*

*«Ischemic stroke occurs because of a loss of blood supply to part of the brain, initiating the ischemic cascade. » [10] In the case of subsequent strokes, it is possible to mitigate consequences if physicians can determine stroke etiology. The paper describes particular approach how to apply Artificial Intelligence for purposes of separating two major acute ischemic stroke (AIS) etiology subtypes: cardiac and large artery atherosclerosis. [1] Four deep neural network architectures and simple ensemble method are used in the approach.*

## 1. Introduction

Stroke is the second-leading cause of death worldwide despite of significant development of medical science and practice. As in other areas of medicine early and accurate diagnosis plays significante role in the fight against disease.

Nowadays mechanical thrombectomy has become the standard of care treatment for acute ischemic stroke from large vessel occlusion. As a result, retrieved clots became amenable to analysis. Healthcare professionals are currently attempting to apply deep learning-based methods to predict ischemic stroke etiology and clot origin. However, unique data formats, image file sizes, as well as the number of available pathology slides create challenges which requires specific approaches to solve them [1].

In such terms simple classification task using transfer learning with SOTA computer vision architectures EfficientNet CNN [6] and Swin Transformer [7] along appropriate data preparation is a suitable solution.

## 2. Dataset

The "train" dataset for the competition contains 754 high-resolution whole-slide digital pathology images in TIF format. Every image represents a blood clot of a patient suffered from an acute ischemic stroke. [1] Also, each image belongs to a one of 632 patients from 11 medicine centers. Each patient may have up to five images and labeled by either CE (Cardioembolic) or LAA (Large Artery Atherosclerosis).

"Other" dataset includes 396 high-resolution whole-slide digital images in TIF format. Each image belongs to a one of 336 patients with no relation to medicine centers. Each patient may have up to five images and labeled by either "Other" or "Unknown".

All images presented in RGB format. Distributions of disk image size and width/height of images in pixels are presented on figure 1.

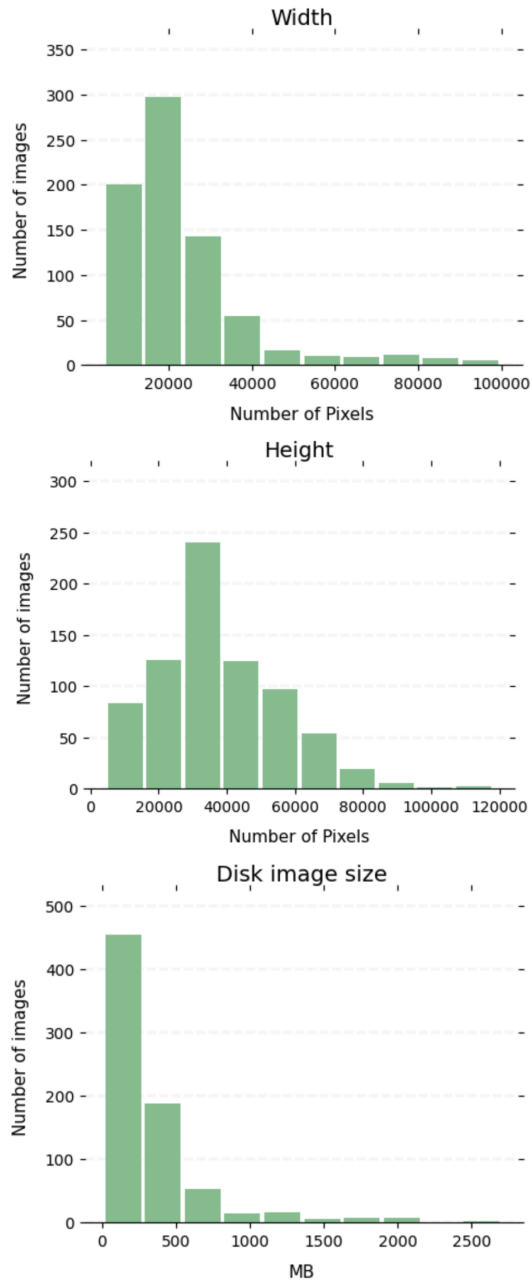

Figure 1. Distributions of disk image size and width/height of images.

The ratio between number of images in the classes CA and LAA for "train" dataset is presented on figure 2.

| label | count | class_rate |
|---|---|---|
| CE | 547 | 0.73 |
| LAA | 207 | 0.27 |

Figure 2. Distributions of disk image size and width/height of images.

## 3. Data preprocessing

Training dataset includes very few high-resolution images in TIF format and that is why it is associated with the following major disadvantages of using original images:
- particular images are stored in files sized up to 2.5 GB, which makes it very hard to work with such files especially if a lot of RAM is not available;
- dataset with high-resolution images is not useful for classification task because of even building SOTA solution of a classification task requires images with resolution in the range from 224x224 pixels to 512x512 pixels;
- original images can have vast 'empty' spaces that doesn't carry any useful information;
- small training sample in deep learning tends models to low ability of generalization and increased risk of overfitting.

To avoid some disadvantages and mitigate the impact of others data preparation includes the following steps:
- pruning and rotation [13]: pruning allows to delete empty columns and rows for a particular image either if height of the image is bigger than its width than the image made mirror turn relative the main diagonal (figure 3 shows original image and figure 4 shows image after pruning and rotation);
- resizing: resizing leads all images to one standard size 1024x1024 which makes further use of images easier due to less disk space and faster transformation operations in RAM;
- augmentation: several following methods of Albumentations library method were used: ToGray, Transpose, VerticalFlip, HorizontalFlip, RandomBrightness, RandomContrast, OneOf(MotionBlur, MedianBlur, GaussianBlur, GaussNoise), OneOf(OpticalDistortion, GridDistortion, ElasticTransform), CLAHE, HueSaturationValue, ShiftScaleRotate, RandomResizedCrop,

Cutout which all have supplemented effect for regularization purposes.

Data normalization with mean = (0.485, 0.456, 0.406), std = (0.229, 0.224, 0.225) is applied for tensor of each image as a last step. Complex data preprocessing described in the section makes possible to train models on available data expecting higher robustness.

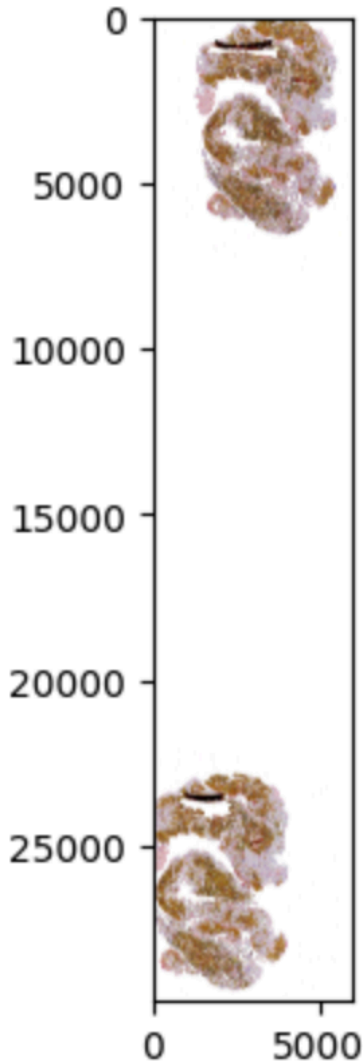

Figure 3. Original Image.

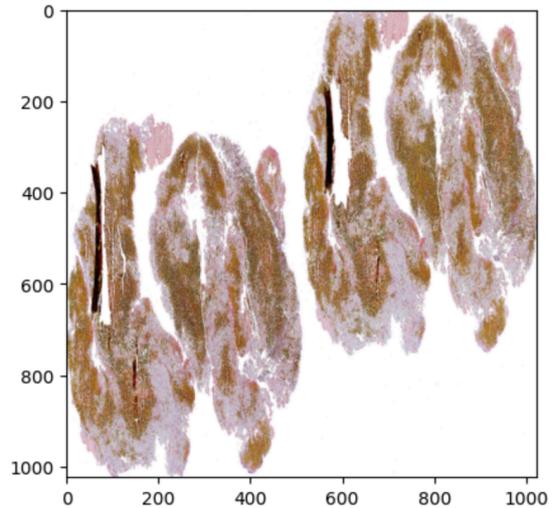

Figure 4. Image after pruning and rotation.

Let's note, that one example in "train" dataset is a patient and a patient can have up to five images. Test performed on the "train" dataset shows that if only last chronological image is taken for each patient, the highest quality of the solution can be achieved. Therefore, as for training purposes as for inference purposes for each patient only last chronological image is taken.

## 4. Models

Classification task of medical images with weak signals seems very challenging notably due to very small "train" dataset. Nevertheless, using a combination of several deep learning architectures helps to avoid their individual disadvantages and to extract signals to separate classes as right as possible.

Several deep learning architectures are used to build final solution: EfficientNet B0 [6] with Noisy Student weights [5] for resolution 512x512, EfficientNet B4 [6] with Noisy Student weights [5] for resolution 384x384, EfficientNet B4 [6] with Noisy Student weights [5] for resolution 512x512, EfficientNet B5 [6] with Noisy Student weights [5] for resolution 512x512, Swin Large with window parameter 7 for resolution 224x224 [7], Swin Large with window parameter 12 for resolution 384x384 [7].

The following successive changes in each model architecture are made:
- liner layer with 128 outputs is added;

- dropout 0.1 is added;
- linear layer with 64 outputs is added;
- liner layer with 2 outputs is added.

Softmax function [12] is used to transform outputs of each model to probabilities.

## 5. Training

We use weighted multi-class logarithmic loss (WMCLL) as evaluating metric.

$$WMCLL = -\frac{\sum_{i=1}^{M} w_i \times \sum_{j=1}^{N_i} \frac{y_{ij}}{N_i} \times \ln p_{ij}}{\sum_{i=1}^{M} w_i}$$

where $N$ is the number of images in the class set, $M$ is the number of classes, $w_i$ – weight for class $i$ (equality of weights has been given), $ln$ is the natural logarithm, $y_{ij}$ is 1 if observation $j$ belongs to class $i$ and 0 otherwise, $p_{ij}$ is the predicted probability that image $j$ belongs to class $i$ [1].

Since evaluating metric is a loss it is used as custom loss function in neural networks training to ensure better convergence in optimization process comparing to standard loss function applied for classification tasks. Therefore, training with minimizing loss function the same time directly minimizes evaluation metric.

Due to the task seems as multilabel classification with two classes, Label Smoothing technique was also used with coefficient 0.01.

All models are trained with Adam optimizer and weight_decay parameter equal to 1e-06. Scheduler ReduceLROnPlateau is used with maximum learning rate 1e-04 and minimum learning rate 1e-05, patience parameter equals 1 and factor to reduce learning rate is 0.1. Maximum number of epochs to train is set to 30 with stopping after six epochs in a row with no decrease of validation loss. Best weights are saved from the best epoch.

For training purposes "train" dataset is splitted into five folds to use it in 5-fold cross-validation scheme. In the first EfficientNet B0 with Noisy Student weights for resolution 512x512 is trained on "train" dataset and after CE an LAA classes are predicted for "other" dataset. To improve model's prediction power images from "other" dataset with probability of classes (CA and LAA) not less than 0.9 are selected and added to "train" dataset with appropriate labels of classes (CA and LAA). Therefore, train folds became bigger. After that EfficientNet B4 with Noisy Student weights for resolution 384x384, EfficientNet B4 with Noisy Student weights for resolution 512x512, EfficientNet B5 with Noisy Student weights for resolution 512x512, Swin Large with window parameter 7 for resolution 224x224, Swin Large with window parameter 12 for resolution 384x384 are also trained on the expanded "train" dataset with 5-folds CV scheme in the training approach described in the article above.

## 6. Results

Final solution is a simple ensemble of five following models: EfficientNet B4 with Noisy Student weights for resolution 384x384, EfficientNet B4 with Noisy Student weights for resolution 512x512, EfficientNet B5 with Noisy Student weights for resolution 512x512, Swin Large with window parameter 7 for resolution 224x224, Swin Large with window parameter 12 for resolution 384x384. Probability of classes CE and LAA calculated as the mean of appropriate probabilities of the five models. In despite of simplicity the ensembling approach among other things allows to increase robustness of prediction. Finally, WCLL = 0.69682 on the public leaderboard and WCLL = 0.67188 on the private leaderboard.

## 7. Conclusions

The paper presents artificial intelligence-based etiology classification of the blood clot origins in ischemic stroke in the situation of the very small data (particularly few hundred images). To allow AI machine work well some strong data augmentations are used among deep CNN (EfficientNet with Noisy Student self-training) and Visual Transformers (Swin) pretrained on ImageNet. The approach shows quite well results on private leaderboard with entering gold zone. It is possible to improve the results by more accurate hyperparameter tuning, more thorough data preprocessing, applying stacking, using TTA (test time augmentation) and using stacking with GBM built on neural networks embeddings.